\pdfoutput=1

\documentclass[11pt]{article}

\usepackage[]{acl2023}

\usepackage{times}
\usepackage{latexsym}
\usepackage{pgfplots}

\usepackage[T1]{fontenc}

\usepackage[utf8]{inputenc}

\usepackage{microtype}

\usepackage{inconsolata}

\usepackage{subcaption}
\usepackage{url}
\usepackage{amssymb}
\usepackage{amsmath}
\usepackage{graphicx}
\usepackage{graphics}
\usepackage{booktabs}
\usepackage{multirow}
\usepackage{color}
\usepackage{enumitem}
\usepackage{makecell}
\usepackage{comment}
\usepackage{nicefrac}
\usepackage{amsfonts}
\usepackage{bm}
\usepackage{bbm}
\usepackage{amsthm}
\usepackage{enumitem}
\setlist[itemize,enumerate]{leftmargin=*}
\usepackage{tikz}
\usepackage{colortbl}
\usepackage{todonotes}
\usepackage{xcolor, soul}

\usepackage{enumitem}
\usepackage{longtable}
\usepackage{arydshln}
\usepackage{CJKutf8}

\definecolor{SRC}{HTML}{A0B1BA}
\definecolor{SRCREF}{HTML}{d3f8e2}
\definecolor{REF}{HTML}{a9def9}

\colorlet{SRC}{SRC!30}
\colorlet{SRCREF}{SRCREF!50}
\colorlet{REF}{REF!50}

\definecolor{plotcolor1}{HTML}{AF58BA}
\definecolor{plotcolor2}{HTML}{00CD6C}
\definecolor{plotcolor3}{HTML}{FFC61E}
\definecolor{plotcolor4}{HTML}{009ADE}

\colorlet{plotcolor1}{plotcolor1!90}
\colorlet{plotcolor2}{plotcolor2!90}
\colorlet{plotcolor3}{plotcolor3!90}
\colorlet{plotcolor4}{plotcolor4!90}

\DeclareRobustCommand{\hlsrc}[1]{{\sethlcolor{SRC}\hl{#1}}}
\DeclareRobustCommand{\hlsrcref}[1]{{\sethlcolor{SRCREF}\hl{#1}}}
\DeclareRobustCommand{\hlref}[1]{{\sethlcolor{REF}\hl{#1}}}

\def\SRC{\hlsrc{\textsc{{\,Src\,}}}}
\def\SRCREF{\hlsrcref{\textsc{Src+Ref}}}
\def\REF{\hlref{\textsc{Ref}}}

\makeatletter
\def\adl@drawiv#1#2#3{%
        \hskip.5\tabcolsep
        \xleaders#3{#2.5\@tempdimb #1{1}#2.5\@tempdimb}%
                #2\z@ plus1fil minus1fil\relax
        \hskip.5\tabcolsep}
\newcommand{\cdashlinelr}[1]{%
  \noalign{\vskip 1.3pt
           \global\let\@dashdrawstore\adl@draw
           \global\let\adl@draw\adl@drawiv}
  \cdashline{#1}[.4pt/2pt]
  \noalign{\global\let\adl@draw\@dashdrawstore
           \vskip 1.3pt}}
\makeatother
\usepackage{footmisc}

\newcommand{\zh}[1]{\begin{CJK}{UTF8}{gbsn}#1\end{CJK}}

\title{The Inside Story: Towards Better Understanding \\ of Machine Translation Neural Evaluation Metrics}

\usepackage[misc]{ifsym}

\author{
    Ricardo Rei\thanks{~~Equal contribution. Corresponding author: \Letter \, \url{ricardo.rei@unbabel.com}}\,\,$^{1,2,4}$,
    Nuno M. Guerreiro$^*$$^{3,4}$,
    Marcos Treviso$^{3,4}$,
    \\
    \bf
    Alon Lavie$^{1}$,
    Luisa Coheur$^{2,4}$,
    André F. T. Martins$^{1,3,4}$
    \\
    $^{1}$Unbabel, Lisbon, Portugal, \,\ $^{2}$INESC-ID, Lisbon, Portugal \\
    $^{3}$Instituto de Telecomunicações, Lisbon, Portugal \\
    $^{4}$Instituto Superior Técnico, University of Lisbon, Portugal
}

\begin{document}
\maketitle
\begin{abstract}
Neural metrics for machine translation evaluation, such as \textsc{Comet}, exhibit significant improvements in their correlation with human judgments compared to traditional metrics based on lexical overlap, such as \textsc{Bleu}. 
Yet neural metrics are, to a great extent, ``black boxes'' that return a single sentence-level score without transparency about the decision-making process. 
In this work, we develop and compare several neural explainability methods and demonstrate their effectiveness for interpreting state-of-the-art fine-tuned neural metrics. 
Our study reveals that these metrics leverage token-level information that can be directly attributed to translation errors, as assessed through comparison of token-level neural saliency maps with \textit{Multidimensional Quality Metrics} (MQM) annotations and with synthetically-generated critical translation errors. To ease future research, we release our code at \url{https://github.com/Unbabel/COMET/tree/explainable-metrics}.


\end{abstract}


\section{Introduction}

Reference-based neural metrics for machine translation evaluation are achieving evergrowing success, demonstrating superior results over traditional lexical overlap-based metrics, such as \textsc{Bleu}~\citep{papineni-etal-2002-bleu} and \textsc{chrF}~\citep{popovic-2015-chrf}, in terms of both their correlation with human ratings and their robustness across diverse domains~\citep{callison-burch-etal-2006-evaluating, smith-etal-2016-climbing, mathur-etal-2020-tangled, kocmi-etal-2021-ship, Freitag-EtAl:2022:WMT}. However, lexical overlap-based metrics remain popular for evaluating the performance and progress of translation systems and algorithms. Concerns regarding trust and interpretability may help explain this~\citep{Leuteretal2022}: contrary to traditional metrics, neural metrics are considered ``black boxes'' as they often use increasingly large models (e.g., the winning metric of the WMT 22 Metrics shared task was a 10B parameter model~\citep{Freitag-EtAl:2022:WMT}). 


\begin{figure}[t]
\centering
\includegraphics[width=\columnwidth]{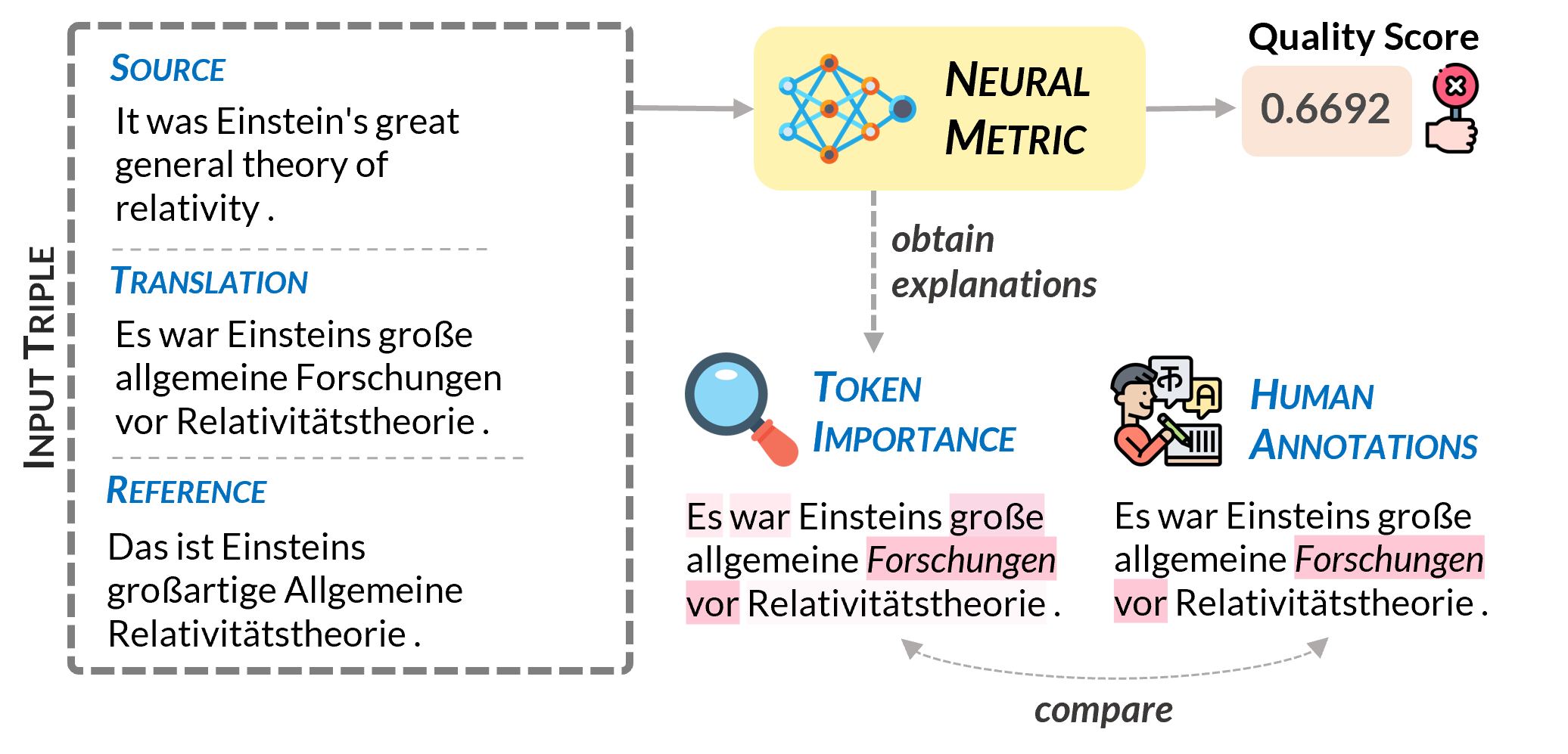}
\caption{Illustration of our approach. In this example, the metric assigns the translation a low score. We aim to better understand this sentence-level assessment by examining the correspondence between our token-level explanations and human annotated error spans.}
\label{fig:paper_procedure}
\end{figure}

While some recent work has focus on explaining the predictions made by \textit{reference-free} quality estimation (QE) systems~\citep{fomicheva-etal-2021-eval4nlp, zerva-EtAl:2022:WMT}, explaining \textit{reference-based} metrics has remained a largely overlooked problem~\citep{Leuteretal2022}. It is an open question whether the observations from studies of explainable QE carry over to this scenario.
Thus, in this work, we fill that gap by turning to state-of-the-art reference-based metrics---we aim to interpret their decision-making process by exploiting the fact that these metrics show consistently good correlations with \textit{Multidimentional Quality Metrics} (MQM)~\cite{freitag-etal-2021-results, Freitag-EtAl:2022:WMT, indicEval-etal-2022}, which are fine-grained quality assessments that result from experts identifying error spans in translation outputs~\cite{lommel2014multidimensional}. 
We hypothesize that reference-based metrics leverage this token-level information to produce sentence-level scores. 
To test this hypothesis, we assess whether our explanations -- measures of token-level importance obtained via attribution and input attribution methods such as attention weights and gradient scores~\citep{treviso-etal-2021-ist, rei-etal-2022-cometkiwi} -- align with human-annotated spans~\citep{fomicheva-etal-2021-eval4nlp, fomicheva-etal-2022-translation, zerva-EtAl:2022:WMT}, as illustrated in Figure~\ref{fig:paper_procedure}. 

Our analysis focuses on two main vectors: (i)~understanding the impact of the reference information on the quality of the explanations; and (ii)~finding whether the explanations can help to identify potential weaknesses in the metrics.
Our main contributions are:

\begin{itemize}[leftmargin=0.35cm, topsep=0.2ex]
    \item We provide a comparison between multiple explainability methods for different metrics on all types of evaluation: \texttt{src}-only, \texttt{ref}-only, and \texttt{src+ref} joint evaluation.
    \item We find that explanations are related to the underlying metric architecture, and that leveraging reference information improves the explanations.
    \item We show that explanations for critical translation errors can reveal weaknesses in the metrics.
\end{itemize}

\section{Explaining Neural Metrics}
\label{sec:exp-metrics}
We aim to explain sentence-level quality assessments of reference-based metrics by producing token-level explanations that align to translation errors. In what follows, we describe the metrics and how we produce the explanations that we study.


\subsection{Metrics}
\label{sec:metrics}

We focus our analysis on two state-of-the-art neural metrics: \textsc{Comet}~\cite{rei-etal-2020-comet} and \textsc{UniTE}~\cite{wan-etal-2022-unite}.\footnote{Ensembles composed of these two metrics were respectively ranked second and third in WMT 2022 Metrics shared task. The winner of WMT 2022 Metrics task~---~\textsc{MetricXXL}~---~is not publicly available~\cite{Freitag-EtAl:2022:WMT}.} While both metrics use a multilingual encoder model based on XLM-~R~\citep{conneau-etal-2020-unsupervised}, they employ distinct strategies to obtain sentence-level quality scores. On the one hand, \textsc{Comet} \textbf{\textit{separately}} encodes the source, translation and reference to obtain their respective sentence embeddings; these embeddings are then combined to compute a quality score. On the other, \textsc{UniTE} \textbf{\textit{jointly}} encodes the sentences to compute a contextualized representation that is subsequently used to compute the quality score. Interestingly, \textsc{UniTE} is trained to obtain quality scores for different input combinations: \texttt{[mt; src]} (\SRC), \texttt{[mt; ref]} (\REF), and \texttt{[mt; src; ref]} (\SRCREF). In fact, when the input is \SRC, \textsc{UniTE} works like TransQuest~\citep{Ranasinghe2020TransQuestTQ}; \REF, like \textsc{Bleurt}~\citep{sellam-etal-2020-bleurt}; and \SRCREF, like \textsc{RoBleurt}~\citep{wan-etal-2021-robleurt}.

\subsection{Explanations via Attribution Methods}
\label{explain-methods}
In this work, we produce explanations using attribution methods that assign a scalar value to each translation token (i.e. a token-level attribution) to represent its importance. While many input attribution methods exist and have been extensively studied in the literature~\citep{ribeiro-etal-2016-trust, pmlr-v70-shrikumar17a, pmlr-v70-sundararajan17a, jain-wallace-2019-attention, atanasova-etal-2020-diagnostic, zamanandbelinkov2022}, we focus specifically on those that have been demonstrated to be effective for explaining the predictions of QE models~\citep{treviso-etal-2021-ist, fomicheva-etal-2022-translation, fernandesetal2022scaffold, zerva-EtAl:2022:WMT} and extend them to our reference-based scenario. Concretely, we use the following techniques to extract explanations:\footnote{For all attention-based methods, we ensemble the explanations from the top 5 heads as this has shown to improve performance consistently over selecting just the best head~\citep{treviso-etal-2021-ist, rei-etal-2022-cometkiwi}. Moreover, we use the full attention matrix, instead of relying only on cross attention information.\label{attn-ensemble-footnote}} 
 \begin{itemize}[leftmargin=0.35cm]
     \item \textbf{embed--align:} the maximum cosine similarity between each translation token embedding and the reference and/or source token embeddings~\citep{hw-tsc:wmt22};
     \item \textbf{grad} $\ell_2$: the $\ell_2$-norm of gradients with respect to the word embeddings of the translation tokens~\citep{arras-etal-2019-evaluating};
     \item \textbf{attention}: the attention weights of the translation tokens for each attention head of the encoder~\citep{treviso-etal-2021-ist};
     \item \textbf{attn} $\times$ \textbf{grad}: the attention weights of each head scaled by the $\ell_2$-norm of the gradients of the value vectors of that head~\citep{rei-etal-2022-cometkiwi}.
 \end{itemize}

\begin{table*}[ht!]
\renewcommand{\arraystretch}{.9}
\small
\centering
\begin{tabular}{clc@{\ \ \ }cc@{\ \ \ }cc@{\ \ \ }cc@{\ \ \ }c}\toprule
\multirow{2}{*}{\rotatebox{0}{{\textsc{Metric}}}}  & \textsc{Explainability} & \multicolumn{2}{c}{ en$\to$de} & \multicolumn{2}{c}{zh$\to$en}  & \multicolumn{2}{c}{en$\to$ru} & \multicolumn{2}{c}{Avg.} \\
& \textsc{Method} & AUC & R@K & AUC & R@K  & AUC & R@K & AUC & R@K \\\midrule
\multicolumn{10}{c}{\textit{\texttt{src}-only$^\star$ evaluation}}\\\cdashlinelr{1-10}
\multirow{4}{*}{\rotatebox{0}{{\fontsize{9}{9}\selectfont \makecell{\textsc{UniTE} \\ \SRC}}}} & embed--align$^{\text{[mt, src]}}$  & 0.587 & 0.339	& \textbf{0.644}	& 0.281	& 0.583	& 0.167	& 0.604	& 0.262 \\
 & grad $\ell_2$ & 0.572	& 0.293	& 0.535	& 0.200	& \textbf{0.620}	& 0.169	& 0.576	& 0.221 \\
 & attention & 0.636	& 0.322	& 0.612	& 0.253	& 0.612	& 0.189	& 0.620	& 0.254 \\
 &  attn $\times$ grad & \textbf{0.707} & \textbf{0.376} & 0.639 & \textbf{0.294} & 0.633 & \textbf{0.211} & \textbf{0.660} & \textbf{0.294} \\\midrule
 \multicolumn{10}{c}{\textit{\texttt{ref}-only evaluation}}\\\cdashlinelr{1-10}
 \multirow{4}{*}{{{\fontsize{9}{9}\selectfont \makecell{\textsc{UniTE} \\ \REF}}}} & embed--align$^{\text{[mt, ref]}}$ & 0.658 & 0.396  & 0.667  & 0.328  & 0.635  & 0.218  & 0.653 & 0.314 \\
 & grad $\ell_2$ & 0.596	& 0.319	& 0.571	& 0.260	& 0.661	& 0.202	& 0.609	& 0.261 \\
 & attention & 0.637 & 0.344 & \textbf{0.670} & 0.335 & 0.652 & 0.224 & 0.653 & 0.301\\
 &  attn $\times$ grad & \textbf{0.725} & \textbf{0.425} & 0.667 & \textbf{0.380} & \textbf{0.660} & \textbf{0.248} & \textbf{0.684} & \textbf{0.351} \\\midrule
\multicolumn{10}{c}{\textit{\texttt{src,ref} joint evaluation}}\\\cdashlinelr{1-10}
\multirow{4}{*}{{\fontsize{9}{9}\selectfont \makecell{\textsc{UniTE} \\ \SRCREF}}} & 
    embed--align$^{\text{[mt, src; ref]}}$  & 0.650	& 0.383	& 0.670	& 0.330	& 0.618	& 0.213	& 0.646	& 0.309  \\ 
 & grad $\ell_2$ & 0.595	& 0.325	& 0.579	& 0.257	& 0.643	& 0.191	& 0.606	& 0.257 \\
 & attention & 0.657	& \textbf{0.421}	& 0.670	& \textbf{0.383}	& 0.649	& 0.223	& 0.659	& 0.342 \\
 &  attn $\times$ grad & \textbf{0.736} & \textbf{0.421} & \textbf{0.674} & \textbf{0.383} & \textbf{0.671} & \textbf{0.248} & \textbf{0.693} & \textbf{0.351} \\\midrule
\multirow{6}{*}{\rotatebox{0}{{\fontsize{9}{9}\selectfont 
\textsc{Comet}}}} & embed--align$^{\text{[mt, src]}}$ & 0.590 & 0.371 & 0.674	& 0.314 & 0.577 & 0.220 & 0.614 & 0.301 \\
& embed--align$^{\text{[mt, ref]}}$  & {0.694} & \textbf{0.425} & 0.696 & 0.355 & {0.647} & 0.275 & \textbf{0.679} & \textbf{0.352} \\
& embed--align$^{\text{[mt, src; ref]}}$ & 0.688 & 0.416 & \textbf{0.697} & \textbf{0.357} & 0.622 & \textbf{0.279} & 0.669 & 0.350 \\
 & grad $\ell_2$ & 0.603 & 0.312 & 0.540 & 0.252 & 0.604 & 0.185 & 0.582 & 0.250 \\
 & attention & 0.604 & 0.351 & 0.592 & 0.259 & 0.633 & 0.209 & 0.608 & 0.268 \\
 &  attn $\times$ grad & \textbf{0.710} & 0.365 & 0.633 & 0.278 & \textbf{0.662} & 0.244 & 0.669 & 0.295 \\\bottomrule
\end{tabular}
\caption{AUC and Recall@K of explanations obtained via different attribution methods for \textsc{Comet} and \textsc{UniTE} models on the MQM data. $^\star$Although \textsc{UniTE} \SRC\,is a \textit{\texttt{src}-only evaluation} metric, it was trained with reference information~\cite{wan-etal-2022-unite}.\vspace{-5pt}}
\label{tab:auc-recall}
\end{table*}

\section{Experimental Setting}
\label{sec:exp_setting}

\paragraph{MQM annotations.} We use MQM annotations from the WMT 2021 Metrics shared task~\cite{freitag-etal-2021-results},\footnote{\url{https://github.com/google/wmt-mqm-human-evaluation}} covering three language pairs --- English-German (en$\to$de), English-Russian (en$\to$ru), and Chinese-English (zh$\to$en)~---in two different domains: News and TED Talks. For each incorrect translation, human experts marked the corresponding error spans. In our framework, these error spans should align with the words that the attribution methods assign higher importance to.



\paragraph{Models.} For \textsc{Comet}, we use the latest publicly available model: \texttt{wmt22-comet-da}~\cite{rei-etal-2022-comet}.\footnote{\url{https://huggingface.co/Unbabel/wmt22-comet-da}} For \textsc{UniTE}, we train our own model using the same data used to train \textsc{Comet} in order to have a comparable setup\footnote{Our implementation differs from the original work by \citet{wan-etal-2022-unite}, See Appendix \ref{app:model_details} for full details.}. We provide full details (training data, correlations with human annotations, and hyperparameters) in Appendix~\ref{app:model_details}. Overall, the resulting 
reference-based \textsc{UniTE} models (\REF\, and \SRCREF) are on par with \textsc{Comet}.


\paragraph{Evaluation.}
We want our explanations to be directly attributed to the annotated error spans, in the style of an error detection task. Thus, we report Area Under Curve (AUC) and Recall@K.\footnote{In this setup, Recall@K is the proportion of words with the highest attribution that correspond to translation errors against the total number of errors in the annotated error span.} These metrics have been used as the main metrics in previous works on explainable QE~\cite{fomicheva-etal-2021-eval4nlp, fomicheva-etal-2022-translation, zerva-EtAl:2022:WMT}.



\section{Results}

\subsection{High-level analysis} 

\paragraph{Explanations are tightly related to the underlying metric architecture.} The results in Table~\ref{tab:auc-recall} show that the predictive power of the attribution methods differ between \textsc{UniTE} and \textsc{Comet}:~attn~$\times$~grad is the best method for \textsc{UniTE}-based models, while embed--align works best for \textsc{Comet}.\footnote{In Appendix~\ref{app:comet_xlmr}, we provide a comparison between the explanations obtained via embed--align with \textsc{Comet} and with its pretrained encoder model, XLM-R.} This is expected as \textsc{UniTE} constructs a joint representation for the input sentences,
thus allowing attention to flow across them; \textsc{Comet}, in contrast, encodes the sentences separately, so it relies heavily on the separate contextualized embeddings that are subsequently combined 
via element-wise operations such as multiplication and absolute difference. 
Interestingly, embed--align and {attn}~$\times$~{grad} were the winning explainability approaches of the WMT 2022 Shared-Task on Quality Estimation~\citep{zerva-EtAl:2022:WMT}. This suggests that explainability methods developed for QE systems can translate well to reference-based metrics. We provide examples of explanations in Appendix~\ref{app:examples}.

\paragraph{Reference information boosts explainability power.} Table~\ref{tab:auc-recall} also shows that, across all metrics, using reference information brings substantial improvements over using only the source information. Moreover, while reference-based attributions significantly outperform source-based attributions, combining the source and reference information to obtain token-level attributions does not consistently yield superior results over using the reference alone. Notably, the best attribution method for \textsc{Comet} does not require any source information. This is interesting: in some cases, reference-based metrics may largely ignore source information, relying heavily on the reference instead.

\subsection{How do the explanations fare for critical translation errors?}
\label{sec:CED}
The MQM data analyzed until now consists primarily of high quality translations, with the majority of annotated errors being non-critical. However, it is important to assess whether our explanations can be accurately attributed to critical errors, as this may reveal potential metric shortcomings. To this end, we employ SMAUG~\cite{SMAUG:WMT22}\footnote{\url{https://github.com/Unbabel/smaug}}, a tool designed to generate synthetic data for stress-testing metrics, to create corrupted translations that contain critical errors. Concretely, we generate translations with the following pathologies: negation errors, hallucinations via insertions, named entity errors, and errors in numbers.\footnote{We corrupt fully correct translations that are not an exact copy of the reference translation. Moreover, as the full suit of SMAUG transformations can only be applied to English data, we focus solely on zh$\to$en translations. Overall, the synthetic dataset consists of 2610 translations. Full statistics about the corrupted data and examples are shown in Appendix~\ref{app:smaug}.} 

\begin{figure}[t]
\centering
\begin{tikzpicture}
\usetikzlibrary{patterns}
\usetikzlibrary{patterns.meta}

\begin{axis}[
set layers=standard,
legend cell align=left,
every axis plot post/.style={/pgf/number format/fixed},
ybar=2.75pt,
bar width=10pt,
x=1.5cm,
y=2.25cm,
ymin=0,
xtick=data,
ylabel=Recall@K,
ylabel style={font=\footnotesize},
x tick label style={align=center, text width=2cm, font=\footnotesize},
y tick label style={font=\footnotesize},
symbolic x coords={\textsc{Neg}, \textsc{Hall}, \textsc{NE}, \textsc{Num}},
visualization depends on=rawy\as\rawy, 
grid = both,
grid style={dashed,gray!40},
axis lines*=left,
clip=true,
area legend,
legend style={at={(0.45, 1.6)},
legend style={draw=gray!70},
legend style={cells={align=center}, font=\small},
anchor=north,legend columns=-1},
legend style={/tikz/every even column/.append style={column sep=0.5cm}},
enlarge x limits={abs=1cm}
]
\addplot[pattern={Lines[angle=0, line width=0.5mm]}, pattern color=plotcolor1!40] coordinates {(\textsc{Neg},0.517) (\textsc{Hall},0.639) (\textsc{NE}, 0.549) (\textsc{Num}, 0.315)};
\addplot[pattern={Lines[angle=45, line width=0.5mm]}, pattern color=plotcolor3!90] coordinates {(\textsc{Neg},0.600) (\textsc{Hall},0.662) (\textsc{NE}, 0.609) (\textsc{Num}, 0.412)};
\addplot[pattern=grid, pattern color=plotcolor2!90] coordinates {(\textsc{Neg},0.606) (\textsc{Hall},0.651) (\textsc{NE}, 0.620) (\textsc{Num}, 0.690)};
\legend{\textsc{Comet}, \textsc{UniTE} \\ \REF, \textsc{UniTE} \\ \SRCREF}
\end{axis}
\end{tikzpicture}
\vspace{-16pt}\caption{Performance of the best attribution methods for \textsc{Comet}, \textsc{Unite} \REF\, and \textsc{UniTE} \SRCREF\, in terms of Recall@K on translations with critical errors: negations (\textsc{Neg}), hallucinations (\textsc{Hall}), named entity errors (\textsc{NE}), and errors in numbers (\textsc{Num}).\vspace{-5pt}}
\label{fig:smaug-comet-unite}
\end{figure}
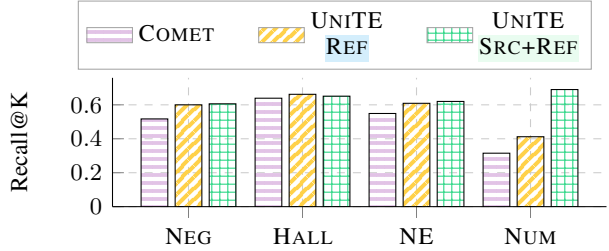

\paragraph{Explanations identify critical errors more easily than non-critical errors.}  Figure~\ref{fig:smaug-comet-unite} shows that explanations are more effective in identifying critical errors compared to other non-critical errors~(see Table~\ref{tab:auc-recall}). Specifically, we find significant performance improvements up to nearly 30\% in Recall@K for certain critical errors. Overall, hallucinations are the easiest errors to identify across all neural metrics. This suggests that neural metrics appropriately identify and penalize hallucinated translations, which aligns with the findings of \citet{guerreiro-etal-2023-looking}. Moreover, explanations for both \textsc{UniTE} models behave similarly for all errors except numbers, where the source information plays a key role in improving the explanations. Notably, contrary to what we observed for data with non-critical errors, \textsc{Comet} explanations are less effective than those of \textsc{Unite} \REF\, and \textsc{UniTE} \SRCREF\, for identifying critical errors.


\paragraph{Explanations can reveal potential metric weaknesses.} Figure~\ref{fig:smaug-comet-unite} suggests that \textsc{Comet} explanations struggle to identify localized errors (negation errors, named entity errors and discrepancies in numbers). We hypothesize that this behavior is  related to the underlying architecture. Unlike \textsc{UniTE}-based metrics, \textsc{Comet} does not rely on soft alignments via attention between the sentences in the encoding process. This process may be key to identify local misalignments during the encoding process. In fact, the attention-based attributions for \textsc{UniTE} metrics can more easily identify these errors. \textsc{Comet}, however, encodes the sentences separately, which may result in grammatical features~(e.g. numbers) being encoded similarly across sentences~\citep{chi-etal-2020-finding, changetal2022}. As such, explanations obtained via embedding alignments will not properly identify these misalignments on similar features. Importantly, these findings align with observations made in~\cite{amrhein-sennrich-2022-identifying, salted_raunak2022}. This showcases how explanations can be used to diagnose and reveal shortcomings of neural-based metrics.

\section{Conclusions and Future Work}
In this paper, we investigated the use of explainability methods to better understand widely used neural metrics for machine translation evaluation, such as \textsc{Comet} and \textsc{UniTE}. Concretely, we analyzed how explanations are impacted by the reference information, and how they can be used to reveal weaknesses of these metrics. Our analysis shows that the quality of the explanations is tightly related to the underlying metric architecture. Interestingly, we also provide evidence that neural metrics like \textsc{Comet} may heavily rely on reference information over source information. Additionally, we show that explanations can be used to reveal reference-based metrics weaknesses such as failing to severely penalize localized critical errors. This opens up promising opportunities for future research on leveraging explanations to diagnose reference-based metrics errors. To support these studies, we call for future datasets illustrating critical errors (e.g., challenge sets~\cite{demetr-2022}) 
to be accompanied by annotated error spans.



\section*{Limitations}
We highlight three main limitations of our work.

First, although we have explored gradient-based explanations that take the whole network into consideration and have been shown to be faithful in previous work~\citep{bastingsetal2021}, we do not explicitly explore how \textsc{Comet} combines the sentence representations in the feed-forward that precedes the encoder model to produce the sentence-level score. 


Second, we have shown that combining attention with gradient information results in the best explanations for \textsc{UniTE}-based metrics. However, from a practical standpoint, running inference and extracting the explainability scores simultaneously may be more computationally expensive than other techniques: gradient-based metrics benefit from GPU infrastructure and require storing all gradient information.

Third, we have not explored extracting explanations in low-resource settings. That is because high-quality MQM annotations for such language pairs are not yet available. Nevertheless, further research in those settings is needed to access the broader validity of our claims.


\section*{Acknowledgements}
This work was partially supported by the P2020 programs (MAIA, contract 045909), the Portuguese Recovery and Resilience Plan (PRR) through project C645008882-00000055, Center for Responsible AI, by the European Research Council (ERC StG DeepSPIN, 758969), by EU’s Horizon Europe Research and Innovation Actions (UTTER, contract 101070631), and by the Funda\c{c}\~{a}o para a Ci\^{e}ncia e Tecnologia (contracts UIDB/50021/2020 and UIDB/50008/2020).

\bibliography{anthology_used,custom}
\bibliographystyle{acl_natbib}

\clearpage
\newpage
\appendix
\begin{table*}[ht!]
\renewcommand{\arraystretch}{.9}
\small
\centering
\begin{tabular}{clc@{\ \ \ }cc@{\ \ \ }cc@{\ \ \ }cc@{\ \ \ }c}\toprule
\multirow{2}{*}{\rotatebox{0}{{\textsc{Metric}}}}  & \textsc{Explainability} & \multicolumn{2}{c}{ en$\to$de} & \multicolumn{2}{c}{zh$\to$en}  & \multicolumn{2}{c}{en$\to$ru} & \multicolumn{2}{c}{Avg.} \\
& \textsc{Method} & AUC & R@K & AUC & R@K  & AUC & R@K & AUC & R@K \\\midrule
 \multirow{3}{*}{{{\fontsize{9}{9}\selectfont \textsc{Xlm-r}}}} & embed--align$^{\text{[mt, src]}}$ & 0.587 & 0.359 & 0.668 & 0.311 & 0.576 & 0.199 & 0.610 & 0.289  \\
& embed--align$^{\text{[mt, ref]}}$ & 0.671 & 0.405 & 0.689 & 0.345 & 0.634 & 0.244 & 0.664 & 0.331 \\
& embed--align$^{\text{[mt, src; ref]}}$  &  0.666 & 0.395 & 0.690 & 0.347 & 0.616 & 0.242 & 0.657 & 0.328 \\\cdashlinelr{1-10}
\multirow{3}{*}{\rotatebox{0}{{\fontsize{9}{9}\selectfont 
\textsc{Comet}}}} & embed--align$^{\text{[mt, src]}}$ & 0.590 & 0.371 & 0.674	& 0.314 & 0.577 & 0.220 & 0.614 & 0.301 \\
& embed--align$^{\text{[mt, ref]}}$  & \textbf{0.694} & \textbf{0.425} & 0.696 & 0.355 & \textbf{0.647} & 0.275 & \textbf{0.679} & \textbf{0.352} \\
& embed--align$^{\text{[mt, src; ref]}}$ & 0.688 & 0.416 & \textbf{0.697} & \textbf{0.357} & 0.622 & \textbf{0.279} & 0.669 & 0.350 \\\bottomrule
\end{tabular}
\caption{AUC and Recall@K of explanations obtained via alignments for \textsc{Comet} and \textsc{Xlm-r} without any further fine-tuning on human annotations.\vspace{-5pt}}
\label{tab:auc-recall-align}
\end{table*}

\section{Model Details}
In Section \ref{sec:metrics}, we employed the latest publicly available model (\texttt{wmt22-comet-da}) for \textsc{Comet}, which emerged as a top-performing metric in the WMT 2022 Metrics task~\cite{Freitag-EtAl:2022:WMT}. To ensure a comparable setting for \textsc{UniTE}~\cite{wan-etal-2022-unite}, we trained our own model. In doing so, we utilized the same data employed in the development of the \textsc{Comet} model by \citep{rei-etal-2022-comet}, without pretraining any synthetic data, as originally suggested. Additionally, our implementation did not incorporate monotonic regional attention, as our preliminary experiments revealed no discernible benefits from its usage. The hyperparameters used are summarized in Table \ref{tab:hp}, while Table \ref{tab:lps} presents the number of Direct Assessments utilized during training. Furthermore, Table \ref{tab:metric-corr} displays the segment-level correlations with WMT 2021 MQM data for the News and TED domains.

Regarding computational infrastructure, a single NVIDIA A10G GPU with 23GB memory was used. The resulting \textsc{UniTE} model has 565M parameters while \textsc{Comet} has 581M parameters.

\label{app:model_details}
\begin{table}[hb!]
\small
\centering
\begin{tabular}{lcc}
\toprule 
\textbf{Hyperparameter}  & \textsc{UniTE}  & \textsc{Comet} \tabularnewline
\midrule 
Encoder Model  & \multicolumn{2}{c}{XLM-R (large)} \tabularnewline
Optimizer  & \multicolumn{2}{c}{AdamW} \tabularnewline
No. frozen epochs  & \multicolumn{2}{c}{0.3} \tabularnewline
Learning rate (LR)  & \multicolumn{2}{c}{1.5e-05} \tabularnewline
Encoder LR.  & \multicolumn{2}{c}{1.0e-06} \tabularnewline  
Layerwise Decay  & \multicolumn{2}{c}{0.95} \tabularnewline
Batch size  & \multicolumn{2}{c}{16}  \tabularnewline 
Loss function  & \multicolumn{2}{c}{MSE} \tabularnewline
Dropout  & \multicolumn{2}{c}{0.1} \tabularnewline
Hidden sizes  & \multicolumn{2}{c}{[3072, 1024]} \tabularnewline
Embedding layer & \multicolumn{2}{c}{Frozen} \\
FP precision  & \multicolumn{2}{c}{16} \\\cdashline{1-3}[.4pt/1pt]
No. Epochs & 1 & 2 \\
\bottomrule
\end{tabular}
\caption{Hyperparameters used to train \textsc{UniTE} and \textsc{Comet} checkpoints used in this work. The only difference between the two is the number of training epochs due to the fact that, for \textsc{UniTE},  the best validation checkpoint is the first one.}
\label{tab:hp}
\end{table}

\subsection{Output Distribution}
\begin{figure*}[ht!]
\centering
\centering
    \begin{subfigure}{\linewidth}
    \centering
      \includegraphics[width=\linewidth]{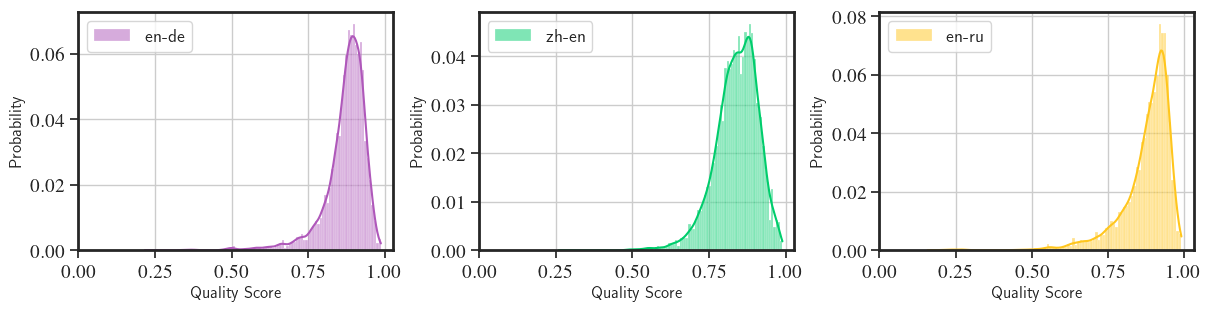}
      \caption{\textsc{Comet}}
    \end{subfigure}\\[1.25ex]
    \begin{subfigure}{\linewidth}
        \centering
      \includegraphics[width=\linewidth]{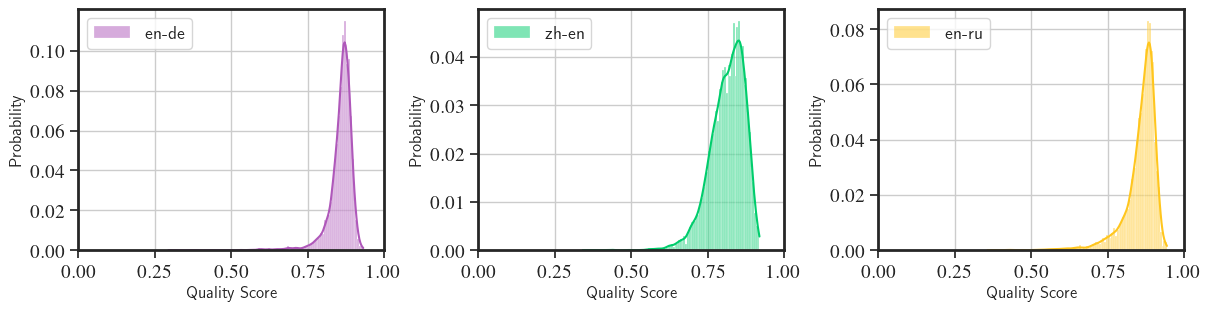}
      \caption{\textsc{UniTE} \SRC}
    \end{subfigure}\\[1.25ex]
    \begin{subfigure}{\linewidth}
        \centering
      \includegraphics[width=\linewidth]{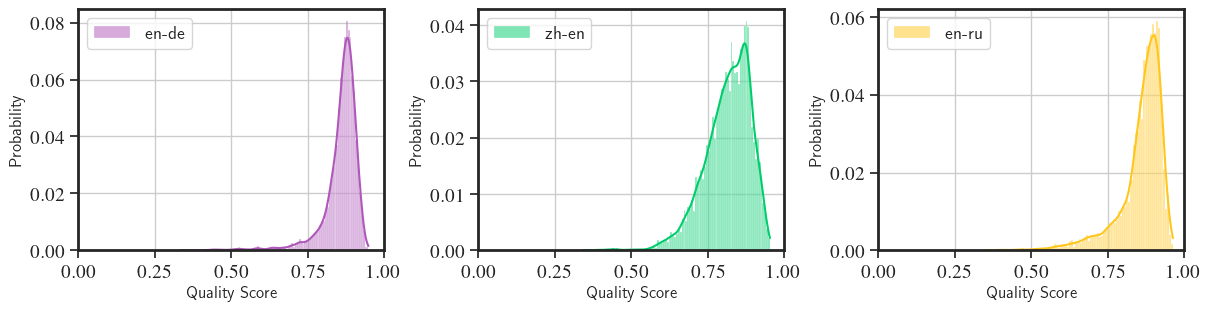}
      \caption{\textsc{UniTE} \REF}
    \end{subfigure}\\[1.25ex]    
    \begin{subfigure}{\linewidth}
        \centering
      \includegraphics[width=\linewidth]{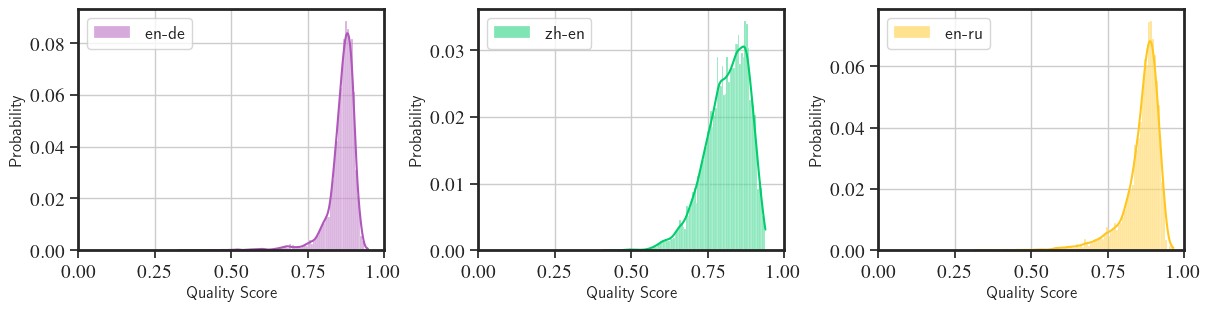}
      \caption{\textsc{UniTE} \SRCREF}
    \end{subfigure}    
\caption{Distribution of scores for all metrics obtained on the MQM data (for all language pairs).}
\label{fig:distribution-ende}
\end{figure*}

To better understand the output of the models and what scores are deemed low, we plotted the output distributions for the two models we used in our study. The average score for English$\to$German data is $0.856$ for the \textsc{Comet} model and $0.870$ for the \textsc{UniTE} model we trained. From Figure~\ref{fig:distribution-ende} we can observe the distribution of scores. This means that the $0.6692$ score from the example in Figure~\ref{fig:paper_procedure} corresponds to a low quality output (5th percentile). 

\begin{table}[hb!]
\small
\centering
\begin{tabular}{lr}
\toprule 
\textbf{Language Pair}  & \textsc{Size}\tabularnewline
\midrule 
zh-en & 126947 \tabularnewline
en-de & 121420 \tabularnewline
de-en &  99183 \tabularnewline
en-zh &  90805 \tabularnewline
ru-en &  79280 \tabularnewline
en-ru &  62749 \tabularnewline
en-cs &  60937 \tabularnewline
fi-en &  46145 \tabularnewline
en-fi &  34335 \tabularnewline
tr-en &  30186 \tabularnewline
et-en &  29496 \tabularnewline
cs-en &  27847 \tabularnewline
en-mr &  26000 \tabularnewline
de-cs &  13804 \tabularnewline
en-et &  13376 \tabularnewline
pl-en &  11816 \tabularnewline
en-pl &  10572 \tabularnewline
lt-en &  10315 \tabularnewline
en-ja & 9578 \tabularnewline
gu-en & 9063 \tabularnewline
si-en & 9000 \tabularnewline
ro-en & 9000 \tabularnewline
ne-en & 9000 \tabularnewline
en-lt & 8959 \tabularnewline
ja-en & 8939 \tabularnewline
en-kk & 8219 \tabularnewline
en-ta & 7890 \tabularnewline
ta-en & 7577 \tabularnewline
en-gu & 6924 \tabularnewline
kk-en & 6789 \tabularnewline
de-fr & 6691 \tabularnewline
en-lv & 5810 \tabularnewline
en-tr & 5171 \tabularnewline
km-en & 4722 \tabularnewline
ps-en & 4611 \tabularnewline
fr-de & 3999 \\\hline
Total & 1027155
\end{tabular}
\caption{Number of direct assessments per language pair used to train \textsc{Comet}~\cite{rei-etal-2022-comet} and the \textsc{UniTE} model used in this work.}  
\label{tab:lps}
\end{table}

\begin{table*}[ht!]
\resizebox{\textwidth}{!}{%
\begin{tabular}{ll|lcccc|cccc}
\toprule 
  \multicolumn{1}{l}{} &
  \multicolumn{1}{l}{} &
  \multicolumn{1}{l}{} &
  \textsc{Bleu} &
  \textsc{chrF} &
  \textsc{YiSi-1} &
  \textsc{Bleurt} &
  \textsc{UniTE} &
  \textsc{UniTE} &
  \textsc{UniTE} &
  \textsc{Comet} \\
    \multicolumn{1}{l}{} &
    \multicolumn{1}{l}{} &
    \multicolumn{1}{l}{} &
     &
     &
     &
     &
    \SRC &
    \REF &
    \SRCREF &
    \texttt{wmt22-comet-da}
    \\\midrule 
\multirow{4}{*}{\rotatebox{90}{{\fontsize{10}{10}\selectfont \textsc{en$\to$de}}}} &
  \multirow{2}{*}{\rotatebox{90}{{\fontsize{10}{10}\selectfont News}}} &
  $\rho$&
  0.077 &
  0.092 &
  0.163 &
  0.307 &
  0.274 &
  \textbf{0.321} &
  0.304 &
  0.297 \\
 &
   &
  $\tau$ &
  0.069 &
  0.092 &
  0.144 &
  0.240 &
  0.222 &
  \textbf{0.248} &
  0.241 &
  0.232 \\ 
 &
   \multirow{2}{*}{\rotatebox{90}{{\fontsize{10}{10}\selectfont TED}}}&
  $\rho$&
  0.151 &
  0.158 &
  0.236 &
  \textbf{0.325} &
  0.311 &
  \textbf{0.335} &
  \textbf{0.338} &
  \textbf{0.329} \\
 &
   &
  $\tau$ &
  0.113 &
  0.146 &
  0.212 &
  0.283 &
  0.264 &
  \textbf{0.301} &
  \textbf{0.298} &
  0.278 \\\cdashline{3-11}[.4pt/1pt]
\multirow{4}{*}{\rotatebox{90}{{\fontsize{10}{10}\selectfont \textsc{en$\to$ru}}}} &
  \multirow{2}{*}{\rotatebox{90}{{\fontsize{10}{10}\selectfont News}}} &
  $\rho$&
  0.153 &
  0.252 &
  0.263 &
  0.359 &
  0.333 &
  \textbf{0.391} &
  \textbf{0.382} &
  0.363 \\
 &
   &
  $\tau$ &
  0.106 &
  0.178 &
  0.216 &
  0.276 &
  0.276 &
  \textbf{0.298} &
  0.297 &
  0.293 \\
 &
   \multirow{2}{*}{\rotatebox{90}{{\fontsize{10}{10}\selectfont TED}}}&
  $\rho$&
  0.154 &
  0.268 &
  0.235 &
  0.286 &
  0.239 &
  0.289 &
  \textbf{0.318} &
  \textbf{0.308} \\
 &
   &
  $\tau$ &
  0.112 &
  0.189 &
  0.204 &
  0.255 &
  0.232 &
  \textbf{0.262} &
  \textbf{0.264} &
  \textbf{0.268} \\\cdashline{3-11}[.4pt/1pt]
\multirow{4}{*}{\rotatebox{90}{{\fontsize{10}{10}\selectfont \textsc{zh$\to$en}}}} &
  \multirow{2}{*}{\rotatebox{90}{{\fontsize{10}{10}\selectfont News}}}&
  $\rho$&
  0.215 &
  0.231 &
  0.301 &
  0.428 &
  0.413 &
  0.438 &
  0.426 &
  \textbf{0.445} \\
 &
   &
  $\tau$ &
  0.165 &
  0.188 &
  0.289 &
  0.341 &
  0.331 &
  0.358 &
  0.352 &
  \textbf{0.371}  \\
 &
   \multirow{2}{*}{\rotatebox{90}{{\fontsize{10}{10}\selectfont TED}}} &
  $\rho$&
  0.155 &
  0.181 &
  0.287 &
  0.295 &
  0.244 &
  0.301 &
  \textbf{0.310} &
  \textbf{0.307} \\
 &
   &
  $\tau$ &
  0.113 &
  0.144 &
  0.216 &
  0.246 &
  0.224 &
  0.265 &
  \textbf{0.266} &
  \textbf{0.269} \\ \bottomrule
\end{tabular}%
}
\caption{Segment-level correlations for WMT 2021 MQM annotations over News and TED domains~\cite{freitag-etal-2021-results}. The metrics are Pearson ($\rho$) and Kendall Tau ($\tau$). Results in bold indicate which metrics are top-performing for that specific language pair, domain and metric according to Perm-Both hypothesis test~\cite{deutsch-etal-2021-statistical}, using 500 re-sampling runs, and setting $p = 0.05$.}
\label{tab:metric-corr}
\end{table*}

\subsection{SMAUG Corpus}
\label{app:smaug}
As we have seen in Section \ref{sec:CED}, we have created synthetic translation errors for the following pathologies: negation errors, hallucinations via insertions, named entity errors, and errors in numbers. Table \ref{tab:smaug} presents a summary of the examples created using SMAUG and in Table \ref{tab:smaug-examples} we show examples of each error category.

\begin{table}[hb!]
\small
\centering
\begin{tabular}{lr}
\toprule 
\textbf{Error Type}  & \textsc{Num Examples}\tabularnewline
\midrule 
\textsc{NE} & 978 \tabularnewline
\textsc{Neg} & 669 \tabularnewline
\textsc{Hall} &  530 \tabularnewline
\textsc{Num} &  432 \\
\midrule
Total & 2609 \\
\bottomrule
\end{tabular}
\caption{Number of examples for each category, synthetically-created using SMAUG~\cite{SMAUG:WMT22}.}  
\label{tab:smaug}
\end{table}

\begin{table}[hb!]
\small
\centering
\begin{tabular}{lrr}
\toprule 
\textbf{Language Pair}  & \textsc{Tokens / Sent.}  & \textsc{Errors / Spans} \tabularnewline
\midrule 
en-de & 528704 / 15310 & 25712 / 3567 \\
en-ru & 525938 / 15074  & 17620 / 7172 \\
zh-en & 603258 / 16506  & 43984 / 10042 \\
\bottomrule
\end{tabular}
\caption{Statistics about MQM data from WMT 2021 Metrics task~\cite{freitag-etal-2021-results} used in our experiments.}  
\label{tab:mqm_stats}
\end{table}

\begin{table*}[ht!]
\small
\begin{tabular}{p{6in}}
\toprule
\textbf{Source:}  \\
\zh{格里沃里表示，分析人士对越南所提出的和平倡议给予认可。} \\\cdashlinelr{1-1}
\textbf{Translation:}  \\
Grivory said that analysts recognize the peace initiative proposed by Vietnam. \\\cdashlinelr{1-1}
\textbf{Reference:}  \\
Grigory said that analysts endorse the peace initiative proposed by Vietnam.\\\cdashlinelr{1-1}
\textbf{\textsc{NE} Error:}  \\
Grivory said that analysts recognize the peace initiative proposed by  \colorbox{gray!20}{Russia}. \\\cdashlinelr{1-1}
\\
\textbf{Source:}  \\
\zh{英国的这一决定预计将会使西班牙的旅游业大受影响。} \\\cdashlinelr{1-1}
\textbf{Translation:}  \\
This decision by the United Kingdom is expected to greatly affect Spain's tourism industry. \\\cdashlinelr{1-1}
\textbf{Reference:}  \\
This decision by the UK is expected to have a significant impact on tourism in Spain.\\\cdashlinelr{1-1}
\textbf{\textsc{Neg} Error:}  \\
This decision by the United Kingdom is expected to greatly \colorbox{gray!20}{benefit} Spain's tourism industry.\\\cdashlinelr{1-1}
\\
\textbf{Source:}  \\
\zh{由于疫情，人们开始在互联网上花费更多的时间。”} \\\cdashlinelr{1-1}
\textbf{Translation:}  \\
Because of the epidemic, people are starting to spend more time on the Internet." \\\cdashlinelr{1-1}
\textbf{Reference:}  \\
For reason of the pandemic, people are starting to spend more time on the Internet. ”\\\cdashlinelr{1-1}
\textbf{\textsc{Hall} Error:}  \\
Because \colorbox{gray!20}{we have a lot} of \colorbox{gray!20}{friends around during} the epidemic, people are starting to spend more time on the \colorbox{gray!20}{mobile devices than on the} Internet." \\\cdashlinelr{1-1} 
\\
\textbf{Source:}  \\
\zh{展销区将展至7月29日。} \\\cdashlinelr{1-1}
\textbf{Translation:}  \\
The exhibition and sales area will be open until July 29. \\\cdashlinelr{1-1}
\textbf{Reference:}  \\
The exhibition will last until July 29.\\\cdashlinelr{1-1}
\textbf{\textsc{Num} Error:}  \\
The exhibition and sales area will be open until July \colorbox{gray!20}{2018}\\
\bottomrule
\end{tabular}
\caption{Synthetically-generated critical errors (\colorbox{gray!20}{highlighted in gray}) created with SMAUG~\cite{SMAUG:WMT22} to assess whether our explanations can be accurately attributed to critical errors.}
\label{tab:smaug-examples}
\end{table*}

\section{Comparison between \textsc{Comet} and \textsc{Xlm-r} Alignments}
\label{app:comet_xlmr}
From Table \ref{tab:auc-recall}, it is evident that the alignments between the reference and/or source and the translation yield effective explanations for \textsc{Comet}. This raises the question of how these alignments compare to the underlying encoder of \textsc{Comet} before the fine-tuning process with human annotations. To investigate this, we examine the results for XLM-R without any fine-tuning, as presented in Table \ref{tab:auc-recall-align}.

Overall, the explanations derived from the alignments of \textsc{Comet} prove to be more predictive of error spans than those obtained from XLM-R alignments. This suggests that during the fine-tuning phase, COMET models modify the underlying XLM-R representations to achieve better alignment with translation errors.

\section{Examples}
\label{app:examples}
In Tables \ref{tab:several-examples-ende} and \ref{tab:several-examples-zh-en}, we show examples of \textsc{Comet} explanations for Chinese$\to$English and English$\to$German language pairs, respectively. We highlight in \colorbox{gray!20}{gray} the corresponding MQM annotation performed by an expert linguist and we sort the examples from highest to lowest \textsc{Comet} scores. From these examples we can observe the following:
\begin{itemize}
    \item Highlights provided by \textsc{Comet} explanations have a high recall with human annotations. In all examples, subword tokens corresponding to translation errors are highlighted in red but we often see that not everything is incorrect.
    \item Explanations are consistent with scores. For example, in the third example from Table \ref{tab:several-examples-zh-en}, the red highlights do not correspond to errors and in fact the translation only has a major error \colorbox{gray!20}{griffen}. Nonetheless, the score assigned by \textsc{Comet} is a low score of $0.68$ which is faithful to the explanations that was given even if the assessment does not agree with human experts.
\end{itemize}

\begin{table*}[ht!]
\small
\begin{tabular}{p{6in}}
\toprule
\textbf{Source:}  \\
And yet, the universe is not a silent movie because the universe isn't silent. \\\cdashlinelr{1-1}
\textbf{Translation:}  \\
Und dennoch ist das Universum kein Stummfilm, weil das Universum nicht \colorbox{gray!20}{still} ist. \\\cdashlinelr{1-1}
\textbf{\textsc{Comet} score: $0.8595$} \\
\textbf{\textsc{Comet} explanation: } \\
\colorbox{red!8}{\_Und} \colorbox{red!14}{\_dennoch} \colorbox{red!1}{\_ist} \colorbox{red!1}{\_das} \colorbox{red!1}{\_Univers} \colorbox{red!1}{um} \colorbox{red!1}{\_kein} \colorbox{red!1}{\_Stu} \colorbox{red!1}{mm} \colorbox{red!1}{film} \colorbox{red!2}{,} \colorbox{red!11}{\_weil} \colorbox{red!0}{\_das} \colorbox{red!1}{\_Univers} \colorbox{red!1}{um} \colorbox{red!2}{\_nicht} \colorbox{red!17}{\_still} \colorbox{red!3}{\_ist} \colorbox{red!0}{.} \\
\\
\textbf{Source:}  \\
And yet black holes may be heard even if they're not seen, and that's because they bang on space-time like a drum. \\\cdashlinelr{1-1}
\textbf{Translation:}  \\
Und dennoch \colorbox{gray!20}{werden} Schwarze Löcher \colorbox{gray!20}{vielleicht gehört}, auch wenn sie nicht gesehen werden, und das liegt daran, dass sie wie eine Trommel auf die Raumzeit schlagen. \\\cdashlinelr{1-1}
\textbf{\textsc{Comet} score: $0.7150$} \\
\textbf{\textsc{Comet} explanation: } \\
\colorbox{red!0}{\_Und} \colorbox{red!0}{\_dennoch} \colorbox{red!11}{\_werden} \colorbox{red!0}{\_Schwarz} \colorbox{red!1}{e} \colorbox{red!0}{\_Lö} \colorbox{red!1}{cher} \colorbox{red!17}{\_vielleicht} \colorbox{red!8}{\_gehört} \colorbox{red!1}{,} \colorbox{red!1}{\_auch} \colorbox{red!1}{\_wenn} \colorbox{red!1}{\_sie} \colorbox{red!11}{\_nicht} \colorbox{red!12}{\_gesehen} \colorbox{red!10}{\_werden} \colorbox{red!2}{,} \colorbox{red!8}{\_und} \colorbox{red!14}{\_das} \colorbox{red!11}{\_liegt} \colorbox{red!14}{\_daran} \colorbox{red!0}{,} \colorbox{red!1}{\_dass} \colorbox{red!1}{\_sie} \colorbox{red!3}{\_wie} \colorbox{red!7}{\_eine} \colorbox{red!3}{\_Tro} \colorbox{red!4}{mmel} \colorbox{red!4}{\_auf} \colorbox{red!6}{\_die} \colorbox{red!1}{\_Raum} \colorbox{red!1}{zeit} \colorbox{red!15}{schlagen} \colorbox{red!0}{.} \\
\\
\textbf{Source:}  \\
Ash O'Brien and husband Jarett Kelley say they were grabbing a bite to eat at Dusty Rhodes dog park in San Diego on Thursday, with their three-month-old pug in tow. \\\cdashlinelr{1-1}
\textbf{Translation:}  \\
Ash O'Brien und Ehemann Jarett Kelley sagen, dass sie am Donnerstag im Hundepark Dusty Rhodes in San Diego einen Happen zu essen \colorbox{gray!20}{griffen}, mit ihrem drei Monate alten Mops im Schlepptau. \\\cdashlinelr{1-1}
\textbf{\textsc{Comet} score: $0.6835$} \\
\textbf{\textsc{Comet} explanation: } \\
\colorbox{red!0}{\_Ash} \colorbox{red!0}{\_O} \colorbox{red!8}{'} \colorbox{red!0}{Bri} \colorbox{red!0}{en} \colorbox{red!1}{\_und} \colorbox{red!14}{\_Ehe} \colorbox{red!7}{mann} \colorbox{red!1}{\_Ja} \colorbox{red!1}{rett} \colorbox{red!16}{\_Kel} \colorbox{red!16}{ley} \colorbox{red!7}{\_sagen} \colorbox{red!0}{,} \colorbox{red!10}{\_dass} \colorbox{red!3}{\_sie} \colorbox{red!3}{\_am} \colorbox{red!1}{\_Donnerstag} \colorbox{red!4}{\_im} \colorbox{red!3}{\_Hunde} \colorbox{red!3}{park} \colorbox{red!3}{\_Du} \colorbox{red!4}{sty} \colorbox{red!4}{\_Rhod} \colorbox{red!0}{es} \colorbox{red!3}{\_in} \colorbox{red!2}{\_San} \colorbox{red!2}{\_Diego} \colorbox{red!9}{\_einen} \colorbox{red!24}{\_Happ} \colorbox{red!13}{en} \colorbox{red!18}{\_zu} \colorbox{red!11}{\_essen} \colorbox{red!1}{\_} \colorbox{red!27}{griff} \colorbox{red!9}{en} \colorbox{red!10}{\_} \colorbox{red!15}{,} \colorbox{red!7}{\_mit} \colorbox{red!2}{\_ihrem} \colorbox{red!2}{\_drei} \colorbox{red!2}{\_Monate} \colorbox{red!2}{\_alten} \colorbox{red!18}{\_M} \colorbox{red!19}{ops} \colorbox{red!9}{\_im} \colorbox{red!23}{\_Schle} \colorbox{red!27}{ppt} \colorbox{red!21}{au} \colorbox{red!0}{.} \\
\\
\textbf{Source:}  \\
It was Einstein's great general theory of relativity. \\\cdashlinelr{1-1}
\textbf{Translation:}  \\
Es war Einsteins große allgemeine \colorbox{gray!20}{Forschungen vor} Relativitätstheorie. \\\cdashlinelr{1-1}
\textbf{\textsc{Comet} score: $0.6692$} \\
\textbf{\textsc{Comet} explanation: } \\
\colorbox{red!9}{\_Es} \colorbox{red!8}{\_war} \colorbox{red!1}{\_Einstein} \colorbox{red!3}{s} \colorbox{red!12}{\_große} \colorbox{red!8}{\_allgemein} \colorbox{red!7}{e} \colorbox{red!19}{\_Forschung} \colorbox{red!10}{en} \colorbox{red!22}{\_vor} \colorbox{red!2}{\_Relativ} \colorbox{red!3}{ität} \colorbox{red!0}{s} \colorbox{red!2}{the} \colorbox{red!3}{ori} \colorbox{red!5}{e} \colorbox{red!0}{.} \\
\\
\textbf{Source:} \\
There's mask-shaming and then there's full on assault. \\\cdashlinelr{1-1}
\textbf{Translation:}  \\
Es gibt \colorbox{gray!20}{Maskenschämen} und dann \colorbox{gray!20}{ist} es  \colorbox{gray!20}{voll bei} Angriff.
\\\cdashlinelr{1-1}
\textbf{\textsc{Comet} score: $0.2318$} \\
\textbf{\textsc{Comet} explanation: } \\
\colorbox{red!12}{\_Es} \colorbox{red!16}{\_gibt} \colorbox{red!4}{\_Mask} \colorbox{red!14}{en} \colorbox{red!22}{schä} \colorbox{red!16}{men} \colorbox{red!12}{\_und} \colorbox{red!19}{\_dann} \colorbox{red!10}{\_ist} \colorbox{red!10}{\_es} \colorbox{red!22}{\_voll} \colorbox{red!19}{\_bei} \colorbox{red!8}{\_Angriff} \colorbox{red!1}{\_} \colorbox{red!13}{.} \\
\bottomrule
\end{tabular}
\caption{Saliency map for \textsc{Comet} explanation scores for a set of en$\to$de examples. Comparing the token-level explanations with the MQM annotation (\colorbox{gray!20}{highlighted in gray}) reveals the source of correspondence between specific token-level translation errors and the resulting scores.}
\label{tab:several-examples-ende}
\end{table*}

\begin{table*}[ht!]
\small
\begin{tabular}{p{6in}}
\toprule
\textbf{Source:}  \\
\zh{我想告诉大家 宇宙有着自己的配乐， 而宇宙自身正在不停地播放着。 因为太空可以想鼓一样振动。} \\\cdashlinelr{1-1}
\textbf{Translation:}  \\
I want to tell you that the universe has its own \colorbox{gray!20}{iconic} soundtrack and the universe itself is \colorbox{gray!20}{constantly} playing non-stop because space can vibrate like a drum. \\\cdashlinelr{1-1}
\textbf{\textsc{Comet} score:} $0.8634$ \\
\textbf{\textsc{Comet} explanation: }\\
\colorbox{red!0}{\_I} \colorbox{red!0}{\_want} \colorbox{red!0}{\_to} \colorbox{red!0}{\_tell} \colorbox{red!0}{\_you} \colorbox{red!0}{\_that} \colorbox{red!0}{\_the} \colorbox{red!0}{\_univers} \colorbox{red!0}{e} \colorbox{red!0}{\_has} \colorbox{red!0}{\_its} \colorbox{red!0}{\_own} \colorbox{red!18}{\_icon} \colorbox{red!10}{ic} \colorbox{red!0}{\_soundtrack} \colorbox{red!1}{\_and} \colorbox{red!0}{\_the} \colorbox{red!0}{\_univers} \colorbox{red!4}{e} \colorbox{red!1}{\_itself} \colorbox{red!1}{\_is} \colorbox{red!14}{\_constantly} \colorbox{red!1}{\_playing} \colorbox{red!1}{\_non} \colorbox{red!0}{-} \colorbox{red!0}{stop} \colorbox{red!0}{\_because} \colorbox{red!0}{\_space} \colorbox{red!0}{\_can} \colorbox{red!0}{\_vibra} \colorbox{red!0}{te} \colorbox{red!0}{\_like} \colorbox{red!0}{\_a} \colorbox{red!0}{\_drum} \colorbox{red!0}{.} \\
\\
\textbf{Source:}  \\
\zh{另外,吉克隽逸和刘烨作为运动助理,也围绕运动少年制造了不少爆笑话题。} \\\cdashlinelr{1-1}
\textbf{Translation:}  \\
In addition, as sports assistants, \colorbox{gray!20}{Ji Kejunyi} and Liu Ye have also created a lot of hilarious topics around sports teenagers. \\\cdashlinelr{1-1}
\textbf{\textsc{Comet} score:} $0.8214$ \\
\textbf{\textsc{Comet} explanation: }\\
\colorbox{red!0}{\_In} \colorbox{red!0}{\_addition} \colorbox{red!1}{,} \colorbox{red!3}{\_as} \colorbox{red!1}{\_sports} \colorbox{red!1}{\_assistant} \colorbox{red!2}{s} \colorbox{red!3}{,} \colorbox{red!4}{\_Ji} \colorbox{red!11}{\_Ke} \colorbox{red!16}{ju} \colorbox{red!10}{nyi} \colorbox{red!1}{\_and} \colorbox{red!1}{\_Li} \colorbox{red!1}{u} \colorbox{red!1}{\_Ye} \colorbox{red!11}{\_have} \colorbox{red!0}{\_also} \colorbox{red!0}{\_created} \colorbox{red!0}{\_a} \colorbox{red!0}{\_lot} \colorbox{red!0}{\_of} \colorbox{red!0}{\_} \colorbox{red!0}{hila} \colorbox{red!0}{rious} \colorbox{red!0}{\_topic} \colorbox{red!0}{s} \colorbox{red!0}{\_around} \colorbox{red!0}{\_sports} \colorbox{red!0}{\_teenager} \colorbox{red!0}{s} \colorbox{red!0}{.} \\
\\
\textbf{Source:}  \\
\zh{一番言论让场上的少年和运动领队们都倒吸一口凉气。} \\\cdashlinelr{1-1}
\textbf{Translation:}  \\
The remarks made the teenagers and the sports leaders on the field gasp a \colorbox{gray!20}{sigh of relief}. \\\cdashlinelr{1-1}
\textbf{\textsc{Comet} score:} $0.7793$ \\
\textbf{\textsc{Comet} explanation: }\\
\colorbox{red!8}{\_The} \colorbox{red!1}{\_re} \colorbox{red!1}{marks} \colorbox{red!1}{\_made} \colorbox{red!8}{\_the} \colorbox{red!0}{\_teenager} \colorbox{red!3}{s} \colorbox{red!1}{\_and} \colorbox{red!10}{\_the} \colorbox{red!0}{\_sports} \colorbox{red!0}{\_leaders} \colorbox{red!1}{\_on} \colorbox{red!0}{\_the} \colorbox{red!1}{\_field} \colorbox{red!4}{\_gas} \colorbox{red!7}{p} \colorbox{red!3}{\_a} \colorbox{red!18}{\_sig} \colorbox{red!11}{h} \colorbox{red!13}{\_of} \colorbox{red!20}{\_relief} \colorbox{red!1}{\_} \colorbox{red!12}{.} \\
\\
\textbf{Source:}  \\
\zh{强烈的阳光是如此地刺眼，} \\\cdashlinelr{1-1}
\textbf{Translation:}  \\
The intense sunlight is \colorbox{gray!20}{so harsh;} \\\cdashlinelr{1-1}
\textbf{\textsc{Comet} score:} $0.7561$ \\
\textbf{\textsc{Comet} explanation: }\\
\colorbox{red!0}{\_The} \colorbox{red!10}{\_intense} \colorbox{red!1}{\_sun} \colorbox{red!1}{light} \colorbox{red!2}{\_is} \colorbox{red!2}{\_so} \colorbox{red!15}{\_har} \colorbox{red!13}{sh} \colorbox{red!14}{;} \\
\\
\textbf{Source:}  \\
\zh{如今，我们希望能够 给这部关于宇宙的 宏伟的视觉作品 配上声音。} \\\cdashlinelr{1-1}
\textbf{Translation:}  \\
\colorbox{gray!20}{Today}, we hope to be able \colorbox{gray!20}{to give} this magnificent visual work \colorbox{gray!20}{of} the universe a sound. \\\cdashlinelr{1-1}
\textbf{\textsc{Comet} score:} $0.7073$ \\
\textbf{\textsc{Comet} explanation: }\\
\colorbox{red!11}{\_Today} \colorbox{red!2}{,} \colorbox{red!2}{\_we} \colorbox{red!14}{\_hope} \colorbox{red!4}{\_to} \colorbox{red!11}{\_be} \colorbox{red!16}{\_able} \colorbox{red!3}{\_to} \colorbox{red!13}{\_give} \colorbox{red!2}{\_this} \colorbox{red!2}{\_magnific} \colorbox{red!3}{ent} \colorbox{red!2}{\_visual} \colorbox{red!3}{\_work} \colorbox{red!14}{\_of} \colorbox{red!0}{\_the} \colorbox{red!3}{\_univers} \colorbox{red!10}{e} \colorbox{red!16}{\_a} \colorbox{red!4}{\_sound} \colorbox{red!0}{.} \\
\bottomrule
\end{tabular}
\caption{Saliency map for \textsc{Comet} explanation scores for a set of zh$\to$en examples. Comparing the token-level explanations with the MQM annotation (\colorbox{gray!20}{highlighted in gray}) reveals the source of correspondence between specific token-level translation errors and the resulting scores.}
\label{tab:several-examples-zh-en}
\end{table*}

\end{document}